%% file: main.tex
\documentclass[sigconf]{acmart}

\usepackage{tabularx}
\usepackage{caption}
\usepackage{subcaption}
\usepackage{colortbl}
\usepackage{multirow}
\usepackage{multicol}
\usepackage{xcolor}
\usepackage{hyperref}
\usepackage{diagbox}

\copyrightyear{2021} 
\acmYear{2021} 
\setcopyright{acmlicensed}\acmConference[CODS COMAD 2021]{8th ACM IKDD CODS and 26th COMAD}{January 2--4, 2021}{Bangalore, India}
\acmBooktitle{8th ACM IKDD CODS and 26th COMAD (CODS COMAD 2021), January 2--4, 2021, Bangalore, India}
\acmPrice{15.00}
\acmDOI{10.1145/3430984.3431026}
\acmISBN{978-1-4503-8817-7/21/01}

\begin{document}

\title{Revisiting Low Resource Status of Indian Languages in Machine Translation}

\author{Jerin Philip}
\authornote{Both authors contributed equally to this work.}
\affiliation{%
  \institution{IIIT Hyderabad}
  \streetaddress{Gachibowli}
  \city{Hyderabad}
  \state{Telengana}
  \postcode{500032}}
\email{jerin.philip@research.iiit.ac.in}

\author{Shashank Siripragada}
\authornotemark[1]
\affiliation{%
  \institution{IIIT Hyderabad}
  \streetaddress{Gachibowli}
  \city{Hyderabad}
  \state{Telengana}
  \postcode{500032}}
\email{shashank.siripragada@alumni.iiit.ac.in}

\author{Vinay P. Namboodiri}
\affiliation{%
  \institution{University of Bath}
  \streetaddress{}
  \city{Bath}
  \state{UK}
  \postcode{500032}}
\email{vpn22@bath.ac.uk}

\author{C.V. Jawahar}
\affiliation{%
  \institution{IIIT Hyderabad}
  \streetaddress{Gachibowli}
  \city{Hyderabad}
  \state{Telengana}
  \postcode{500032}}
\email{jawahar@iiit.ac.in}

\renewcommand{\shortauthors}{Philip et al.}

\newcommand{\langdir}[2]{$\textit{#1}{\rightarrow}\textit{#2}$}

\newcommand{\cc}[2]{
    \cellcolor[rgb]{#1,#1,#1}{#2}
}

\newcommand{\vspacehackuniform}{}

\begin{abstract}
Indian language machine translation performance is hampered due to the lack of large scale multi-lingual sentence aligned corpora and robust benchmarks. Through this paper, we provide and analyse an automated framework to obtain such a corpus for Indian language neural machine translation (NMT) systems. Our pipeline consists of a baseline NMT system, a retrieval module, and an alignment module that is used to work with publicly available websites such as press releases by the government. The main contribution towards this effort is to obtain an incremental method that uses the above pipeline to iteratively improve the size of the corpus as well as improve each of the components of our system. Through our work, we also evaluate the design choices such as the choice of pivoting language and the effect of iterative incremental increase in corpus size. Our work in addition to providing an automated framework also results in generating a relatively larger corpus as compared to existing corpora that are available for Indian languages. This corpus helps us obtain substantially improved results on the publicly available WAT evaluation benchmark and other standard evaluation benchmarks.

\end{abstract}


\maketitle

\section{Introduction}

Advances in machine translation, language-modelling, and other natural language-processing has led to a steep increase performance on tasks for many high-resource languages \cite{ott2018,edunov2018understanding}. One major driving factor is many western languages which become test-beds for the methods are already high-resource, which works in favour of methods which are data hungry \cite{koehn2017six}. The high resource European counterparts have supporting projects like Europarl \cite{koehn2005europarl}, Paracrawl \cite{banon-etal-2020-paracrawl}. These have enabled large scale sentence aligned corpora to be developed. Similar efforts have not been realized for languages in the Indian subcontinent~\cite{joshi2020state}. Evidently, attempts need to be undertaken to improve this situation. Our work directly addresses this lacuna in the Indian machine translation setting. Specifically, through this work, we aim to achieve the following objectives:
\begin{itemize}
    \item Provide a large scale sentence aligned corpus in 11 Indian languages, viz. CVIT-PIB corpus that is the largest multilingual corpus available for Indian languages as can be seen from Table~\ref{table:iter0-dataset-stats}.
    \item Demonstrate that such a corpus can be obtained automatically with no human effort using iterative alignment method. 
    \item Provide robust standardized evaluation methodology and strong baselines that can be adopted and improved upon to ensure systematic progress in machine translation in Indian languages.
\end{itemize}

We briefly examine the alternatives to our approach and argue the need for adopting the proposed approach. 

\paragraph{Working at Scale}
There have been impressive works for low-resource languages at scale \cite{aharoni2019massively,lepikhin2020gshard,schwenk2019wikimatrix}, for instance working with 1620 language pairs \cite{schwenk2019wikimatrix} . However, not all these advances are feasible with regard to the compute resources available to standard academic research groups. Specifically, large models that converge faster, transfer more, and improve performance even for low-resource languages \cite{aharoni2019massively,lepikhin2020gshard} are not trainable on hardware available to many research groups. Hence, we argue that this approach is not viable for Indian machine translation research, at this moment.

\paragraph{Presently available corpora and baselines}
Unfortunately, research in Indian language machine translation suffers from the lack of suitable publicly available models and baselines. Those that are available are rather limited in scope or evaluation.
For instance, a widely used corpus that is available is the ILCI Corpus \cite{jha2010tdil}. The corpus has 50K sentences aligned across many languages in the country. However, this corpus is limited to specific domains. The evaluation strategies on this corpora in literature also lacks comparability with no standard test-split. Despite these limitations, the corpus has been used by several reported sources as training data in literature to develop and study Machine Translation
\cite{anthes2010automated,kunchukuttan2014sata,goyal-etal-2020-efficient}. However, this corpus is not useful for applications like \citet{kr2019towards}, due to the limitations. The Workshop on Asian Translation (WAT) \cite{nakazawa-etal-2018-overview,nakazawa2017overview,nakazawa2019overview} on the other hand provides a standardized platform for a few languages. Similarly the Workshop on Machine Translation (WMT) \cite{barrault2019findings} from time to time hosts tasks with directions involving Indian Languages. Unfortunately, though several iterations of these tasks have concluded, to the best of our knowledge, there are no trained models that are publicly available at the moment. We summarize and list the presently available corpora in Table~\ref{table:iter0-dataset-stats}. It is evident that large scale multi-lingual corpora are lacking presently. There are
multiple attempts in the past for developing ML solutions for Indian languages, including   \cite{bhattacharyya-etal-2016-statistical} and other attempts such as~\cite{Singh2019NMTIL,murthy-etal-2019-addressing,goyal-etal-2020-efficient,kunchukuttan2016learning}. Unfortunately, most of these
methods do not evaluate on a standard benchmarks/dataset for reliable comparison of the performance. They also have inferior performance to our approach and do not provide publicly available models. This thereby inhibits research in the community.

\paragraph{Proposed approach}
We believe, that most methods applicable to the high-resource languages should work just as well in Indian languages in presence of the same amount of data. A simple solution which maintains Occam's razor is to change the low-resource situation, as more content is created online in many Indian Languages which is not pursued as much as it should be. {Steps have been taken towards improving the situation in the monolingual corpus space \cite{kunchukuttan2020ai4bharat,kakwani2020indicnlpsuite}}. %

\begin{figure}[h!]
\centering
  \includegraphics[width=0.8\linewidth]{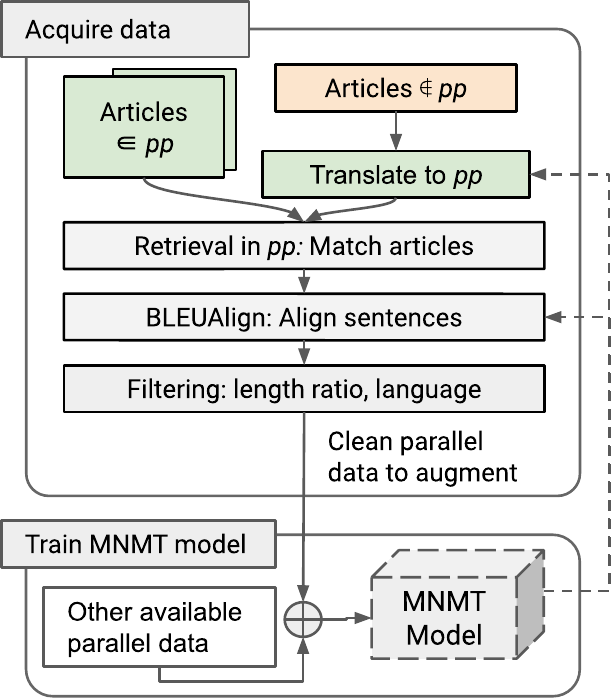}
  \caption{Iterative alignment pipeline used for expanding the corpus for Indian languages. We observe that (i) A better MNMT model leads to better alignment and larger  corpus (ii) Larger corpus leads to better MNMT model. We iterate until no further improvement is observed. The dashed lines indicate application of the trained MNMT model. \textit{pp} stands for an arbitrary pivot language.}
  \label{fig:pipeline}
  \vspacehackuniform{}
\end{figure}

In this work we demonstrate how, using recent advances in Multilingual Neural Machine Translation (MNMT) \cite{vaswani2017attention,johnson2017google,dabre2020comprehensive} in an Expectation Maximization (EM) setup in the face of incomplete data, it is possible to change the status-quo of low resource to produce larger corpus and strong baselines in machine-translation for several Indian languages. A first-step towards this was taken in \citet{siripragada-etal-2020-multilingual} where we presented the CVIT-PIBv0.0 and CVIT Mann Ki Baat corpora. We substantially extend and refine this work through an iterative pipeline illustrated in Figure~\ref{fig:pipeline}. {We also co-opt some ideas proposed in low-resource adaptation for NMT proposed by \cite{sennrich-zhang-2019-revisiting}.} In the process, we attain stronger baselines in translating the involved language-directions. Our contributions summarized are as follows:

\begin{enumerate}
    \item \textbf{Low Resource\textrightarrow High Resource}: We extensively study the iterative-alignment methods provided by \citet{sennrich2011iterative} in the context of CVIT-PIBv0.0 dataset created for \citet{siripragada-etal-2020-multilingual}. Through successful execution of these methods, we increase the corpora size aggregated over all language-pairs from our previous 613K\footnote{In \citet{siripragada-etal-2020-multilingual}, we report this as 408K considering only English alignments, in this work we consider all-directions.} to 2.78M (\textasciitilde{} 353\% increase) that we term the CVIT-PIB corpus v1.3. 
     \item \textbf{Comparable and Strong Baselines}: We report consequent stronger baselines for MNMT in Indian languages from the improvement in data, validating the utility of the corpora we provide. The final MNMT model covers 11 languages and 110 language-directions with competitive or state-of-the-art performance in {12 tasks} on public leaderboards.
    \item \textbf{Trained models and code}: We release the source-code, trained models and the datasets\footnote{\href{http://preon.iiit.ac.in/~jerin/bhasha/}{http://preon.iiit.ac.in/\textasciitilde{}jerin/bhasha/}} to further research in this area and to aide applications that could be enabled by a functional MT. {To the best of our knowledge, these are the only trained models available for translation with focus on Indian languages at the time of writing this document.}
\end{enumerate}

The rest of this document is organized as follows: In Section \ref{section:iterativePIB}, we describe using traditional methods in MT based alignment to improve the parallel corpus across 11 Indian Languages. In Section \ref{section:strongerBaselines} we report stronger baselines for MNMT in Indian languages across many available public tasks.

\section{Iterative Alignment for PIB}
\label{section:iterativePIB}
We apply the methods in \citet{sennrich2011iterative} to iteratively improve parallel data. The procedure is analogous to expectation-maximization (EM) algorithm. In the expectation step, we use noisy alignments of parallel sentences from news articles to get a meaningful signal to obtain a better Maximum Likelihood Estimate (MLE) function for the MT model. In the maximization-step the improved MT model is used to obtain stronger alignments. Unlike \citet{sennrich2011iterative}, we use an MNMT model in place of the Statistical Machine Translation (SMT) model.
In this section, we provide details about the corpus, the methods used to obtain the same and analysis of its characteristics.

\subsection{Data Sources}
To obtain an initial Multilingual NMT~(MNMT) system, we rely on the datasets compiled from several sources listed in Table \ref{table:iter0-dataset-stats}. We use backtranslation \cite{sennrich2016improving} to improve data in Hindi and Telugu.

The Press Information Bureau (PIB) is used in this work as a source for articles published in several Indian Languages to extract a multiparallel corpus. The PIB is very similar to a newspaper publishing in several languages except with strong one-to-one matches between documents and monotonic sentences which provide more parallel sentences through automatic sentence alignment algorithms. %
This section describes using the same crawled content as \citet{siripragada-etal-2020-multilingual} and focuses on improving the quality of alignments, in an attempt to consequently improve corpus size and performance of the MNMT model.

\input{tables/iter0-dataset-stats}

We are aware of the existence of PMIndia \cite{haddow2020pmindia}, a source of similar nature and motivation as PIB, but make a conscious choice not to use it in this work to prevent data-leakage issues of possible overlap with one of our test-sets CVIT Mann Ki Baat~\cite{siripragada-etal-2020-multilingual}.

\subsection{Iterative Alignment}

Our iterative alignment procedure requires document and sentence alignment algorithms, an MNMT formulation which is trained again with refreshed data in each iteration. We describe the constituent components illustrated in Figure~\ref{fig:pipeline} and describe the iterative procedure ahead.

\subsubsection{Text Processing}

\paragraph{Text Cleaning and Standardization} We allow for noise on the web in the pipeline and avoid any linguistic features. There is noise present in documents generated in the past with Indian languages content. 
A desideratum for retrieval and matching is that the model is capable of handling such noise.
This could be unicode issues, non-standard or normalized text which is present all-across sources on the web. Some amount of noise is mitigated by past works \cite{bhattacharyya-etal-2016-statistical} by standardising unicode, scripts etc\footnote{\href{https://github.com/anoopkunchukuttan/indic_nlp_library}{https://github.com/anoopkunchukuttan/indic\_nlp\_library}}.%

\paragraph{Tokenization}
\label{tokenization}
We use SentencePiece~\cite{kudo2018sentencepiece} to tokenize sentences into subword-units. The subwords which cover a corpus are decided optimizing likelihood of a unigram language-model over a large corpus and candidate subwords in EM steps. Recent works \cite{edunov2018understanding,ng2019facebook} addressing high-resource western-languages follow a joint vocabulary of some 32K-64K subwords as a subword model creation hyperparameter. We observed in our early experiments that in the presence of
huge imbalance of data among languages and the huge difference in scripts unlike major European languages, for example, this approach leads to subwords which reduce to characters for the less represented languages. In order to avoid the artifacts from such a subword learning strategy, we instead choose 4K subwords for each of the languages involved and take a union of these to generate the final vocabulary\footnote{This design choice is partially inspired by the reasoning and supporting experiments in \citet{sennrich-zhang-2019-revisiting}.}. The process results in a vocabulary of 40K subword-units tokens for 11 languages which we maintain fixed across all iterations. We note that the artifacts can also be mitigated by a temperature based sampling for sentences among languages as \citet{aharoni2019massively}.

\paragraph{Filtering parallel pairs} To obtain a filtered corpora at every iteration, we allow only sentence-pairs into to the training pipeline where source length to target length ratio is in $[0.5, 2.0]$. We also use~\texttt{langid}\footnote{\url{https://github.com/saffsd/langid.py}}, a language identifier through writing script to filter sentences with foreign language tokens.

\subsubsection{Alignment Algorithms}
 To perform document alignment we translate the articles to a common pivot language.
 We use the translations to rank candidate-matches in the pivot language. %
SentencePiece tokenization of each sentence eliminates requirement of a curated stop-words list, enabling us to compute similarity in the search space of any desired pivot language. Cosine similarity on the \emph{term frequency - inverse document frequency}~(tf-idf)~\cite{buck-koehn-2016-quick} is put to use to rank retrieved articles in the space of pivot language. Search space to find and rank candidate articles matching a translation is restricted to only in a vicinity of dates (2 days) of posted news articles.

Upon obtaining aligned document pairs, we use Bleualign~\cite{sennrich2010mt}, an MT based sentence alignment algorithm. Other conventional sentence length based alignment algorithms such as Gale-Church~\cite{gale-church-1993-program} also exist, but we rely on MT based alignment as the performance of the NMT model increases with every iteration resulting in better sentence alignment. Bleualign also aggressively filters reducing false matches \cite{banon-etal-2020-paracrawl}.

\input{tables/accuracy-size-bleu}

\subsubsection{Multilingual Neural Machine Translation (MNMT) Model} %
We use fairseq~\cite{ott2019fairseq} for training a Transformer-Base~\cite{vaswani2017attention} based MNMT system. The model we begin with is same as our first multilingual model in \citet{siripragada-etal-2020-multilingual}. However, unlike \citet{siripragada-etal-2020-multilingual}, to refine the CVIT-PIBv0.0 dataset further, we choose a many-to-English model formulation trained to translate only from non-English languages to English. This is advantageous because (1) it enables faster training and retraining time, (2) the setting provides more capacity to English decoding which improves translation performance and consequently -- retrieval in English. The model parameters - encoder, decoder and embeddings are shared among all languages. We additionally use tied embeddings \cite{press2017using} at the encoder and decoder. In this work, we denote our first many to many model with no CVIT-PIBv0.0 dataset augmentation as M2M-0, the following many to English models with incremental dataset variations as M2EN-1, M2EN-2, M2EN-3. We additionally consider the best model from \citet{siripragada-etal-2020-multilingual} after training with CVIT-PIBv0.0 dataset augmentation, denoted in this work as M2M-1, to attempt another translation to a non English and possibly related-language as an alternative for retrieval.

\subsubsection{Iterations} 

\begin{figure}[h!]
\includegraphics[width=\linewidth]{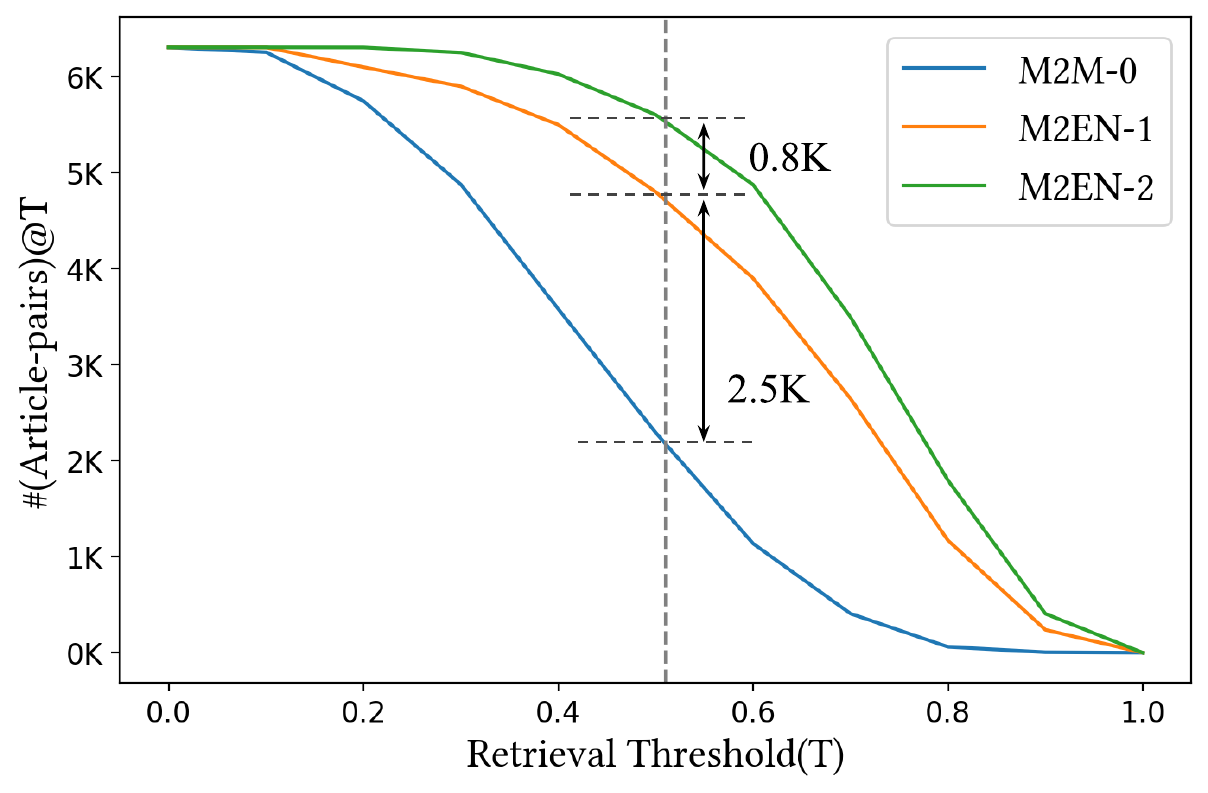}
\caption{Figure illustrating number of articles on the Y-axis obtained at a given threshold on X-axis. We maintain a constant threshold~(dotted line) of~\emph{0.51} across model iterations M2M-0, M2EN-1 and M2EN-2. In case of Marathi,  when compared to M2M-0 we acquire an additional 2.5K article pairs using M2EN-1 and 0.8K more using M2EN-2. From the graph, we observe saturation in articles pairs after iteration 2 indicating a point of diminishing returns.}
\label{fig:marathi-iterations}
\vspacehackuniform{}
\end{figure}

In each iteration, we initialize training with the model from previous iteration~(warm-start). This helps to reduce training time when compared to a model training from scratch~(cold-start) as we benefit from learning in the previous iterations. 
To maintain a constant increment in articles, we set a threshold on keeping a retrieved candidate at a constant value for each language. We observe that the scores improve with successive iterations, consequently obtaining more matching documents. We stop the iteration process at a stage of diminishing returns, i.e there is no prospect of a justifiable increase in corpus size (See Figure \ref{fig:marathi-iterations}). %
Note that in future, further expansion through an increasing number of documents is always possible.

\subsection{Discussions}

The many-languages involved and the disparity in sizes in training lead to a setting where we can dissect and study several aspects. First we study the iterative alignment process, comparing retrieval accuracy, BLEU scores and increase in corpora side by side.

We track BLEU \cite{papineni2002bleu} for the MT model, a \emph{pseudo} retrieval accuracy for the retrieval pipeline and count of successful sentence alignments for increase in corpora over iterations. We employ the following technique to arrive at our \emph{pseudo} retrieval accuracy. The articles which match in dates and ministry information from meta-information are collected. In many languages, there are enough true-positive matches to report the consequent evaluations as a proxy to retrieval accuracy.

These numbers reported in Table \ref{table:accuracy-size-bleu} indicate mostly consistent trends reflecting improvement in BLEU scores in translating to English for the CVIT Mann Ki Baat {(WAT-2020 test split\footnote{\url{http://lotus.kuee.kyoto-u.ac.jp/WAT/indic-multilingual/}}}), retrieval accuracies and consequently the resulting acquired data-sizes. The BLEU scores improve with the addition of more data, while the retrieval accuracy and data sizes improve with updating the parallel-corpus generated from PIB using the higher-performing MNMT model. 
Through successive iterations, the corpora increases in size and gets refined. We observe in Table~\ref{table:accuracy-size-bleu}, most increase for the languages with already good data (Hindi). In three iterations, we add a net 744K sentences aligned to English on top of the 408K (82\% increase) sentences in the previous release, CVIT-PIBv0.0. But however, it is worth noting that some languages which are aided by the transfer and the new data have almost increased an order in sizes in iterations later (Marathi, Oriya and Bengali). 
Once the parallel corpus extraction from PIB is saturated (at M2EN-3), we crawl another 7 months of articles to expand the data further, and run another round of the alignment routine over the entire corpus. We obtain the multiparallel corpus by getting sentence alignments amongst other languages by bridging through the English part of the existing English-centric data. 
The process ends ups providing an additional 2.17M (353\% increase) sentences to the previous 613K sentences aligned across languages, resulting in a corpora size of \emph{2.78M}, viz. CVIT-PIBv1.3\footnote{There exists a CVIT-PIBv0.2 used for WAT-2020 and WMT-2020. See Appendix for more information.}.

In the case of Oriya, we observe no date-based matches. This leads to depending entirely on accurate document pair retrieval for extracting a corpus of reasonable alignment accuracy. The preliminary efforts used to retrieve and align data in \citet{siripragada-etal-2020-multilingual} comprises of Bible~\cite{parida2020odiencorp} which has been observed to transfer poorly to other domains consequently to poor sentence level alignment accuracies for Oriya in our earlier models~\cite{siripragada-etal-2020-multilingual}. However, with every iteration in Tables \ref{table:accuracy-size-bleu} and \ref{table:bleu-benchmarks} we observe increase in BLEU scores when translating to English. This leads to better retrieval and sentence level accuracies. %

In our early experiments, we had left Urdu out from the iterative alignment procedures due to unresolved bugs in the text processing in the pipeline. This lead to no changes in Urdu to English corpus sizes for the first three iterations. However, this paves the way for a case study to evaluate the effect of the improvements in other multilingual language-pairs in translating to and from Urdu as we include Urdu in the performance evaluations. We notice that as the remaining resources improve and better alignments are put in place, and with no further Urdu data enhancements, there are improvements in the BLEU scores of language pairs involving Urdu, observed in Table \ref{table:bleu-benchmarks} {through M2EN-1 to M2EN-3}. %

\input{tables/pib-buffer-viz}

To summarize, we provide a new corpus and a method to expand the corpus from publicly available sources.
\begin{itemize}
    \item We make a new large sentence aligned corpus (CVIT-PIBv1.3)  available for researchers, as a result of the iterative alignment described above. The detailed analysis of the iterative alignment procedure is provided in Table 2. 
    \item This new corpus is possibly the largest sentence aligned multi-lingual corpus for Indian languages as can be observed through Table 1  %
    in number.
    \item Table 3 shows the corresponding increase in size and percentage increase from our previous effort using the same document set as in \cite{siripragada-etal-2020-multilingual} with the increment obtained only through the proposed procedure. We further provide strong baselines that are presented in the next section.
    \item We also hint at a method that can allow continuous expansion of this corpus and eventually enabling a class of recent language processing methods on Indian languages.
\end{itemize}

Having described the process of iterative refining and enriching parallel-corpus resources for Indian Languages, we use the resulting corpora in training two models for obtaining baselines - one many-to-many (M2M) and the other many-to-English (M2EN). Since these are the next iterations, we label these M2M-4 and M2EN-4. In addition to these, we consider a model with 32K output vocabulary giving more vocabulary to English (M2M, M2EN uses 4K) which is denoted by M2EN-4-32K, disabling tied-embeddings.
The {three} models are used to establish strong baselines for the 11 languages and consequent 110 directions involved which we describe in the next section.

\section{Stronger and Repeatable Baselines}
\label{section:strongerBaselines}

It is important to further research to first take stock of where we are, and often simple baselines which compete in comparison to sophisticated methods serve this exact purpose \cite{xiao2018simple,arora2019simple}. We consider the possibility using standard approaches which work for high-resource languages to provide such baselines for translation among Indian languages. Our previous work \citet{philip2019baseline} reports baselines for multilingual translation models for 5 Indian Languages and English. Further improvements are brought about in \citet{siripragada-etal-2020-multilingual} in several tasks adding more languages. It is also important to take care that these are ``simple baselines'' which are comparable \cite{post2012constructing}, leaving plenty of room for improvements ahead.

Addressing simplicity, through the aforementioned works and this one, the model architecture and hence the learner's ``capacity'' is kept constant (Transformer-Base). Without increasing capacity, the experiments continue taking on all available translation tasks in Indian languages. A task here corresponds to language-directions or domains of datasets.
The model we use is expected to be as general purpose as possible, after improving the data situation. There are no linguistics based priors in our methods or explicit handling of noise. This provides room for linguistics based improvements to build on simultaneously raising a call to revisit some older propositions.
Addressing comparability, we comprehensively cover and compare with test-sets in prior-art and the shared tasks.

\input{tables/bleu-benchmarks}

\subsection{On Repeatability of Objective Evaluations}

We address two important aspects (1) standard test-sets available and (2) reproducible evaluations.

\subsubsection{Test Sets}
We identify two class of test-sets among Indian languages, (1) which corresponds to ILCI which many early works evaluated translation quality on and (2) associated with the WAT or WMT tasks, which provide a leaderboard and standardized evaluations for comparison. To cover limitations of these test-sets, we propose a new test-set CVIT Mann Ki Baat in \citet{siripragada-etal-2020-multilingual}. We proceed to summarize how we compare to past work reporting numbers on these test-sets.

\subsubsection{Comparable reports of BLEU Scores} \citet{post2018call} addresses several issues of reproducibility and fair comparisons in reporting BLEU scores. In this work, we make our evaluations consistent with WAT leaderboard and provide a package to reproduce the procedure locally\footnote{\href{https://github.com/jerinphilip/wateval}{https://github.com/jerinphilip/wateval}}. For WMT tasks, we report the values from the portal\footnote{\href{http://matrix.statmt.org/matrix}{http://matrix.statmt.org/matrix}}(gu-en) and the SacreBLEU\footnote{BLEU+case.mixed+numrefs.1+smooth.exp+tok.13a+version.1.4.12} signature (ta-en). We make the hypotheses generated among all test-sets available\footnote{\href{https://github.com/shashanksiripragada/generation-results}{https://github.com/shashanksiripragada/generation-results}} in case of a requirement to re-evaluate and compare with non-BLEU metric.

\subsection{Results and Discussions}

We begin by discussing the general merits of the model especially coming out of the multilingual formulations, and proceeding to elaborate on where we stand with regard to existing literature. For translating in to English, our best M2EN model (M2EN-4-32K) provide a cumulative improvement of +25 BLEU and an average improvement of +3.5 BLEU, compared to the previous best known multilingual model M2M-1~\cite{siripragada-etal-2020-multilingual} {with a similar coverage of languages}, with the same model capacities on CVIT Mann Ki Baat test set in Table~\ref{table:accuracy-size-bleu}. This clearly points to the improvement consequent of change in data situation in the involved Indian languages. In Table \ref{table:mkb-benchmark}, we report BLEU scores of the M2M-4 in a grid indicating the performance in the language-directions the model applies. We also highlight the improvement magnitude in color.

The M2EN model being stronger in the tasks it is trained towards compared to the corresponding M2M model, and is consistent with what is established in literature\cite{aharoni2019massively}. A reasoning for this disparity is not enabling temperature based sampling as in \citet{aharoni2019massively} to balance out all language-pairs and correct for the imbalance in huge number of English aligned sentences existing in the training data. In \citet{bapna2019simple}, corrections for the imbalance are observed to have led to degradation in high-resource languages. Note that both models are trained with the same capacity, and gain in BLEU scores and capability in translating more languages is a major advantage of the M2M model.

To study generalization, we take the models from the iterative procedure described in Section \ref{section:iterativePIB} and evaluate their performance on all available test-sets. The results can be observed in Tables \ref{table:accuracy-size-bleu} and \ref{table:bleu-benchmarks}.

Over the span of a few incremental works, we have significantly improved Hindi to English translation by a margin of +3 BLEU since \citet{philip2019baseline}, obtaining higher numbers by using simple methods. This is unlike many rounds of distillation, hyperparameter optimization done by other groups to reach similar range of values\cite{nakazawa-etal-2018-overview}. The other languages also have similar improvements in all directions.  %

Despite being not trained with the provided Gujarati data and using the data from ILCI and {CVIT-PIBv1.3 corpus}, we are able to achieve a BLEU score of 25.3 and 25.6 with our models M2EN-3 and M2M-3 respectively, competitive to the best BLEU score of 26.9 \cite{li2019niutrans} in WMT 2019 gu-en task. The best performing model did several rounds of backtranslation, distillation and multilingual formulation leveraging Hindi. Among these, we have only taken advantage of multilingual formulation and on top, pure data augmentation at the moment. A caveat here is that we have not put in efforts into filtering the test-data from the training data. But our corpus collection is independent of the news-sources WMT19 used and the Gujarati-English directions are as good as the claim, also supported by the results on CVIT Mann Ki Baat and ILCI test sets.

The BLEU scores on CVIT Mann Ki Baat test-set are provided in Table \ref{table:mkb-benchmark}. We notice the most improvements in Oriya involved directions. Our overall multilingual model seems to have improved at {M2M-4} in comparison to M2M-1. Despite not enforcing any linguistic priors, we get strong performance in many related languages.

It is also worth noting the correlations between Tables \ref{table:pib-final-grid} and \ref{table:mkb-benchmark}, that the highest improvements in BLEU has been for directions involving those languages for which more data has been added. With a model of fixed capacity throughout, simply increasing data has given increase in performance. The trend suggests the possibility to collect more sentence-aligned parallel text as a means to improve performance of machine translation models for Indian languages.

\subsection{Comparison with Previous Works}

\input{tables/Hindi_English_Comparisons.tex}

\input{tables/mkb-benchmark-M2M4}

\paragraph{Non-standard comparisons} Comparison is not standardized since previous methods evaluate on their own test set or on non-standard splits of the ILCI corpus that are not publicly available. This could be common in the initial stages of research for any community. Our work addresses it by evaluating it on more standard benchmarks with clear publicly available  test splits. \citet{kunchukuttan2014sata} attempts to build a collection of {SMT} models covering 11 Indian languages, similar to this work, except training and testing on splits from ILCI corpus (2K test-sentences, 500 for validation and remaining for training). However, the split is not available. \citet{goyal-etal-2020-efficient} once again report numbers on ILCI, using a similar split strategy as \citet{kunchukuttan2014sata}.  \citet{murthy-etal-2019-addressing} compares with a similar test-set of 2K ILCI sentences. Similarly, \citet{goyal-etal-2020-efficient} uses the ILCI test-set with a similar strategy to apply a formulation taking advantage of ``related-languages''. %
In our experiments we have found ILCI to be domain-specific (health, tourism) and providing false perception of high-scores by models which fail to generalize  \cite{siripragada-etal-2020-multilingual}. Due to the lack of reproducibility and comparability and the known demerits, we do not recommend future comparisons on any arbitrary  ILCI test-set for benchmarking general purpose translation systems.

\paragraph{Comparison on Hindi and English} Hindi is the Indian language that has received highest attention with multiple attempts for translatiion to and from English. 
The results for comparison for Hindi-English on publicly available standard benchmarks that can be accessed are provided in Table~\ref{table:hindi_english_comparisons}. We obtain the highest scores for two of the tasks, i.e. IITB and WAT-ILMPC Hindi-English evaluations. Note that the highest BLEU score for IITB English to Hindi was obtained by our previous approach~\cite{philip-etal-2018-cvit}.

\paragraph{Comparison on Public Leader Boards}
We provide detailed results for our method on publicly available leaderboards in Table~\ref{table:bleu-benchmarks} and Table~\ref{table:mkb-benchmark}. These can be used for comparisons and evaluations by various methods. As mentioned previously, these are on WAT tasks, WMT Tasks and CVIT Mann Ki Baat evaluation set. In all these tasks our models perform well obtaining state of the art results for several tasks. For instance, we obtain state of the art BLEU score of 19.68 on OdiEnCorp that is much higher than the previous state of the art of 8.6 and 24.77 BLEU score on WAT-ILMPC for Tamil to English that is  higher than the previous state of the art 24.31.

We thus establish  strong baselines for machine translation for Indian languages. Our multilingual model outperforms the previous works and  even many carefully handcrafted MT systems for specific language pairs.

\section{Conclusions and Future Directions}
In this work, we have made contributions to 
change the low-resource status of several Indian Languages for Machine Translation. Specifically (i) We introduced a large corpus that can enable Deep Machine Translation and associated research for these languages. (ii) Domain of these sentences allow to cover wider topics and practically more useful. More importantly, we established the utility of an algorithm that can help to grow the size starting from this state. More data will lead to better models, that in turn better alignment and more data. The corpus is bound to increase in size as more articles get added to PIB and the tooling in place to collect more sentences.

This work also possibly acts as the first NMT
model dealing in Indian Languages that is publicly available for research. We hope this will spur more research within the research groups, specially in India. To enable the same, our code, models and data splits are made publicly available.
Our corpus is now getting used in the WMT and WAT (International and Asian premier machine translation forums), demonstrating the utility. In addition to the parallel-corpus, we also make access to the crawling tools public - which can be used in the future to create document-level NMT datasets in Indian languages.

A challenging question will be the applicability of this method for online resources that are not really created by explicit translation. We believe, that a solution to this problem may not be too far from here. 
The methods used for search and alignment in this paper can be extended to use newspapers and news specific to a time-window in a weakly supervised setting with minimal human effort to enhance the parallel-corpus further. Using embeddings trained to mine parallel sentences have shown promise for High Resource Languages, which we will incorporate into this pipeline in the future.  The meta-information on the stored PIB articles opens up possibilities to study document translation and active learning problems, left for future work.

Our M2EN models have high BLEU scores which allows for an application of backtranslation \cite{sennrich2016improving,edunov2018understanding} to improve the numbers further in the opposite direction (\langdir{en}{xx}). \citet{kim2020and} reports scenarios where unsupervised NMT methods fail for the low-resource Gujarati-English pair due to limitations, and the enhancement of resources here implores a revisit.

With high-performing NMT systems to English from Indian languages, it is possible to create datasets and corpus for use in downstream tasks and by using the models provided by this work to further the research in Indian Languages, with \citet{kunchukuttan2020ai4bharat} using the \emph{CVIT Mann Ki Baat} to evaluate cross-lingual sentence retrieval being an example.
\begin{acks}

The authors acknowledge and thank the discussion and correspondence with Anoop Kunchukuttan. We acknowledge the past contributions of ILCI, WAT in providing a reasonable seed-corpus which enabled building this work. 
We are grateful for the timely correspondence with WAT Organizers, especially Toshiaki Nakazawa which enabled us to conveniently benchmark with existing numbers. 
We  thank Binu Jasim, for his early contribution to a sentence segmenter which runs to date. We thank Pruthwik Mishra, and Shantipriya Parida for inspecting the Oriya samples.
\end{acks}

\clearpage

\bibliographystyle{ACM-Reference-Format}
\bibliography{main}

\appendix

\input{appendix.tex}

\end{document}

%% file: tables/iter0-dataset-stats.tex
\begin{table}[h]
\centering
{\small 
    \begin{tabular}{lrrr}
    \toprule
        Source & \#pairs & \#lang & type \\
        \midrule
        IITB-en-hi~\cite{kunchukuttan2017iit} & 1.5M & 2 & en-hi \\
        UFAL EnTam~\cite{RaBoMorphologicalProcessing2012}& 170K & 2& en-ta \\
        
        WAT-ILMPC~\cite{nakazawa-etal-2018-overview} & 800K & 7 &  xx-en \\
        ILCI~\cite{jha2010tdil} & 550K & 11&  xx-yy \\
        OdiEnCorp~\cite{parida2020odiencorp} & 27K &  2& en-or \\
          \midrule
        Backtranslated-Hindi & 2.5M & 2& en-hi \\
        Backtranslated-Telugu & 500K &2  & en-te \\
         \midrule
         CVIT Mann Ki Baat\cite{siripragada-etal-2020-multilingual} & 41K & 10&xx-yy\\
         PMIndia-Corpus\cite{haddow2020pmindia} & 728K & 13&xx-yy \\
         CVIT-PIBv0.0\cite{siripragada-etal-2020-multilingual} & 613K&11&xx-yy\\
         {CVIT-PIBv0.2} & {1.17M}& {11}&xx-yy\\
         {\textbf{CVIT-PIBv1.3}} & \textbf{2.78M} & 11 & xx-yy\\
         \bottomrule
    \end{tabular}
}
\caption{Publicly available corpuses for Indian languages. The last group of rows were not used for training. CVIT Mann Ki Baat is used for evaluation purposes only and has overlap with PMIndia Corpus.  All other sources are used for training the multilingual model. xx-yy indicates parallel sentences aligned across multiple languages. Last row is the proposed corpus. %
}

\label{table:iter0-dataset-stats}
    \vspacehackuniform{}
\end{table}

%% file: tables/accuracy-size-bleu.tex
\begin{table*}[h]
\centering
    {\small 
    \begin{tabular}{llrrrrrrrrr}
        \toprule

             & Model  & \multicolumn{9}{c}{Languages} \\
             &  &   hi & ta & ml & mr & gu & te & or & bn & pa\\ 
        \midrule
        \multirow{3}{*}{Retrieval Accuracy} & M2M-0  & 76.77 & 54.12 & 45.63 & 34.05 & 52.52 & 24.06 & - & - & -\\
                                        & M2EN-1 & 86.91 & 71.36 & 69.77 & 47.57 & 63.67 & 62.3 & - & - & -\\  
                                        & M2EN-2 & 92.39 & 80.84 &	80.4 & 51.89 &	75.86& 	70.05 & - & -  & -\\  
                                    
        \midrule
        \multirow{7}{*}{\langdir{xx}{en} BLEU} & M2M-0 & 17.63 & 9.49 & 11.2 & 11.35 & 14.44 & 6.98 & - & 10.82 & - \\
                                    & M2M-1  & 20.11 & 13.38 & 14.89 & 15.89 & 19.60 & 10.02  & - & 14.77 & - \\
                                    & M2EN-1 & 21.29 & 14.41 & 15.59 & 16.73 & 20.04 & 9.25  & - & 15.19 & - \\
                                    & M2EN-2 & 22.17 & 15.25 & 16.92 & 17.64 & 21.27 & 9.93  & - & 16.39 & - \\
                                    & M2EN-3 & 22.00 & 15.43 & 16.98 & 18.02 & 21.28 & 10.05 & - & 16.50 & -\\ 
                                    \cmidrule(lr){2-11}
                                    &  M2EN-4 & 22.55 & 16.48 & 18.35 & 19.35 & 23.08 & 13.62 & - & 17.83 & - \\
                             & M2EN-4-32K & 22.98 & 17.05 & 18.86 & 19.53 & 23.39 & 13.74 & - & 18.06 & -\\
        \midrule
        \multirow{3}{*}{Corpus Size} & M2M-0  & 156.3K & 61.0K & 17.0K & 40.0K & 25.5K & 6.0K & 9.1K & 21.6K & 26.3K \\
                                     & M2EN-1 & 189.2K & 73.7K & 28.7K & 71.6K & 26.3K & 5.3K & 20.3K & 42.4K & 24.6K\\  
                                     & M2EN-2 & 195.2K & 87.1K & 32K & 81.0K & 29.4K & 5.7K & 58.5K & 48.3K & 27.1K \\
        \midrule
                                    & Increment & 38.9K & 26.1K & 15.0K & 40.8K & 3.9K & 0 & 49.4K	& 26.8K	& 0.8K \\

        \bottomrule
    \end{tabular}
    }
    \caption{Incremental improvements in Accuracy, \langdir{xx}{en} BLEU scores on \emph{Mann Ki Baat} {(test-split from WAT-2020 used here to stay comparable)} and Corpus size. We observe increments in retrieval accuracies consistent with increase in BLEU scores. WAT-2020 test split does not contain pa and or, while PIB does.}
    \label{table:accuracy-size-bleu}
    \vspacehackuniform{}
\end{table*}

%% file: tables/pib-buffer-viz.tex
\newcolumntype{Y}{>{\small\raggedleft\arraybackslash}p{0.7cm}}

\begin{table*}[h]
\centering
{
\small 
    \begin{tabular}{YYYYYYYYYYYY}
    \toprule
&en&hi&ta&te&ml&ur&bn&gu&mr&or&pa \\ \midrule
en&&\cellcolor[rgb]{0.770319,0.856209,0.935102}{269594}&\cellcolor[rgb]{0.867266,0.919354,0.967520}{118759}&\cellcolor[rgb]{0.897885,0.939039,0.977363}{44888}&\cellcolor[rgb]{0.919416,0.952818,0.984252}{44986}&\cellcolor[rgb]{0.647290,0.803922,0.892042}{202578}&\cellcolor[rgb]{0.843645,0.903606,0.959646}{93560}&\cellcolor[rgb]{0.907113,0.944944,0.980315}{59739}&\cellcolor[rgb]{0.834787,0.897701,0.956694}{117199}&\cellcolor[rgb]{0.814118,0.883922,0.949804}{98230}&\cellcolor[rgb]{0.834787,0.897701,0.956694}{103296}\\
hi&&&\cellcolor[rgb]{0.897885,0.939039,0.977363}{64936}&\cellcolor[rgb]{0.919416,0.952818,0.984252}{28562}&\cellcolor[rgb]{0.934794,0.962661,0.989173}{27154}&\cellcolor[rgb]{0.796401,0.872111,0.943899}{109946}&\cellcolor[rgb]{0.897885,0.939039,0.977363}{49584}&\cellcolor[rgb]{0.916340,0.950850,0.983268}{41583}&\cellcolor[rgb]{0.885582,0.931165,0.973426}{69167}&\cellcolor[rgb]{0.864314,0.917386,0.966536}{61065}&\cellcolor[rgb]{0.849550,0.907543,0.961615}{75188}\\
ta&&&&\cellcolor[rgb]{0.940946,0.966597,0.991142}{17356}&\cellcolor[rgb]{0.934794,0.962661,0.989173}{23599}&\cellcolor[rgb]{0.888658,0.933133,0.974410}{48872}&\cellcolor[rgb]{0.922491,0.954787,0.985236}{32988}&\cellcolor[rgb]{0.931719,0.960692,0.988189}{29182}&\cellcolor[rgb]{0.904037,0.942976,0.979331}{48527}&\cellcolor[rgb]{0.891734,0.935102,0.975394}{44019}&\cellcolor[rgb]{0.894810,0.937070,0.976378}{46340}\\
te&&&&&\cellcolor[rgb]{0.950173,0.972503,0.994095}{10467}&\cellcolor[rgb]{0.931719,0.960692,0.988189}{21141}&\cellcolor[rgb]{0.937870,0.964629,0.990158}{17604}&\cellcolor[rgb]{0.940946,0.966597,0.991142}{16325}&\cellcolor[rgb]{0.937870,0.964629,0.990158}{18169}&\cellcolor[rgb]{0.950173,0.972503,0.994095}{10462}&\cellcolor[rgb]{0.925567,0.956755,0.986221}{25680}\\
ml&&&&&&\cellcolor[rgb]{0.931719,0.960692,0.988189}{20894}&\cellcolor[rgb]{0.940946,0.966597,0.991142}{18136}&\cellcolor[rgb]{0.944022,0.968566,0.992126}{18234}&\cellcolor[rgb]{0.934794,0.962661,0.989173}{22793}&\cellcolor[rgb]{0.934794,0.962661,0.989173}{19390}&\cellcolor[rgb]{0.931719,0.960692,0.988189}{21960}\\
ur&&&&&&&\cellcolor[rgb]{0.900961,0.941007,0.978347}{39290}&\cellcolor[rgb]{0.919416,0.952818,0.984252}{29914}&\cellcolor[rgb]{0.882507,0.929196,0.972441}{49683}&\cellcolor[rgb]{0.891734,0.935102,0.975394}{43733}&\cellcolor[rgb]{0.885582,0.931165,0.973426}{51817}\\
bn&&&&&&&&\cellcolor[rgb]{0.931719,0.960692,0.988189}{25154}&\cellcolor[rgb]{0.916340,0.950850,0.983268}{34025}&\cellcolor[rgb]{0.922491,0.954787,0.985236}{26449}&\cellcolor[rgb]{0.910188,0.946913,0.981300}{35107}\\
gu&&&&&&&&&\cellcolor[rgb]{0.922491,0.954787,0.985236}{30759}&\cellcolor[rgb]{0.919416,0.952818,0.984252}{27140}&\cellcolor[rgb]{0.910188,0.946913,0.981300}{35555}\\
mr&&&&&&&&&&\cellcolor[rgb]{0.885582,0.931165,0.973426}{46999}&\cellcolor[rgb]{0.885582,0.931165,0.973426}{50411}\\
or&&&&&&&&&&&\cellcolor[rgb]{0.894810,0.937070,0.976378}{43138}\\
\bottomrule
    \end{tabular}
}
\caption{Multilingual shared content across language pairs for CVIT-PIBv1.3. Rows and columns indicate language pairs. The highlights are proportional to the increases in corpora sizes compared to the previous release CVIT-PIBv0.0 \cite{siripragada-etal-2020-multilingual}}
\label{table:pib-final-grid}
    \vspacehackuniform{}
\end{table*}

%% file: tables/bleu-benchmarks.tex
\begin{table*}[h]
\centering
    {\small 
    \begin{tabular}{clrrrrrrrrrrrrrrr}
        \toprule
         direction & Model  & {IITB}  & {UFAL} & {OdiEnCorp} & \multicolumn{6}{c}{WAT-ILMPC} & {WMT19} & \multicolumn{2}{c}{WMT20}\\ \cmidrule(lr){6-11} \cmidrule(lr){13-14}
                   &   & hi  & ta & or & hi & ta & te & ml & ur & bn & gu  & ta (test) & ta (dev) \\ 
        \midrule
        \multirow{4}{*}{\langdir{en}{xx}} & M2M-0 & 19.83 & 6.78 & 4.29 & 19.99 & 10.86 & 16.19 & 7.19 & 13.27 & 9.69 & 6.4 & 3.5 & 4.8  \\
                                          & M2M-1 & 20.52 & 7.31 & 5.26 & 20.39 & 11.63 & 16.63 & 7.92 & 16.55 & 9.58 & 9.5 & 4.2 & 6.0 \\
                                          & M2M-3 & 21.20 & 7.22 & 4.78 & 20.92 & 11.95 & 17.10 & 7.70 & 15.78 & 10.13 & 11.3 & 4.9 & 7.1 \\
                                          & M2M-4 & 21.28 & 7.80 & 5.25 & 20.25 & 10.00 & 15.80 & 6.60 & 16.08 & 9.29 & 12.5 & 5.1 & 7.4 \\

        \midrule
        \multirow{9}{*}{\langdir{xx}{en}} & M2M-0  & 21.94 & 18.64 & 11.05 & 27.99 & 17.82 & 21.63 & 11.97 & 20.57 & 16.67  & 17.9 & 12.9 & 12.7\\
                        & M2M-1  & 22.48 & 19.76 & 10.84 & 28.31 & 18.65 & 22.58 & 12.71 & 21.16 & 16.77 & 23.6 & 14.3 & 13.9\\  
                        & M2M-3  & 23.07 & 19.87 & 12.07 & 28.99 & 19.16 & 23.96 & 12.77 & 21.15 & 17.38 & 25.6 & 15.9 & 15.3 \\
                        & M2M-4  & 22.84 & 19.66 & 12.28 & 27.88 & 18.09 & 22.93 & 12.19 & 21.19 & 16.74 & 25.2 & 16.6 & 15.1 \\
                        \cmidrule(lr){2-14}
                        & M2EN-1 & 23.83  & 23.38 & 13.07 & 31.33 & 21.17 & 25.69 & 14.24 & 23.38 & 18.78 & 22.8 & 15.5 & 15.0\\  
                        & M2EN-2 & 24.65  & 25.32 &  15.62 & 32.88 & 23.19 & 28.11 & 15.68 & 24.53 & 20.03 & 24.5 & 16.6 & 16.3\\  
                        & M2EN-3 & 25.26 & 26.08 & 17.76 & 34.09 & 23.85 & 29.47 & 16.38 & 25.88 & 20.62 & 25.3 & 16.7 & 16.4 \\            
                        & M2EN-4 & 25.01 & 26.49 & 17.41 & 33.73 & 23.35 & 30.14 & 15.87 & 26.38 & 19.89 & 24.6 & 17.2 & 16.6 \\
                        & M2EN-4-32K & 24.63 & 27.40 & 19.68 & 34.44 & 24.77 & 31.44 & 17.17 & 27.79 & 21.36 & 24.2 & 17.2 & 17.1 \\
        \bottomrule
    \end{tabular}
    }
    \caption{We report BLEU scores on available publicly available benchmark tasks for Indian Languages. The results on these benchmarks often have models that are specially tuned for various language pairs. We do observe that we obtain state of the art results on 4 of the language pairs and are competitive to other works that are more specific in most cases. This is despite not being specially tuned for these settings.}  %
    \label{table:bleu-benchmarks}
        \vspacehackuniform{}
\end{table*}

%% file: tables/Hindi_English_Comparisons.tex
\begin{table}[h!]
\centering
{\small 
    \begin{tabular}{lrrrr}
    \toprule
        Work & \multicolumn{2}{c}{IITB-hi-en} & \multicolumn{2}{c}{WAT-ILMPC}\\
        & \langdir{en}{hi} & \langdir{hi}{en} & \langdir{en}{hi}  & \langdir{hi}{en} \\
        \midrule
        SMT~\cite{kunchukuttan2017iit} & 11.75 & 14.49& - & - \\
        NMT~\cite{kunchukuttan2017iit} & 12.23 & 12.83 & - & -\\
        \citet{saini2018neural} & 18.22 & - & - & - \\
        \citet{philip-etal-2018-cvit} & \textbf{21.57} & 20.63 & - & - \\
        \citet{dabre-etal-2018-nicts} & & & \textbf{29.65} & 31.51 \\
        \citet{goyal-sharma-2019-ltrc} & - & 18.64 & - & -\\   
        \citet{philip2019baseline}& 20.17 & 22.62 & 26.25 & 31.55 \\
        \citet{siripragada-etal-2020-multilingual}& 20.52 & 22.48 & 20.3 & 28.3 \\ 

          \midrule
        Proposed Methods & 21.28 & \textbf{25.26} & 20.92 & \textbf{34.44}  \\
         \bottomrule
    \end{tabular}
}
\caption{Comparison with publicly available baselines for English to Hindi and vice versa.}
\vspacehackuniform{}
\vspace{-0.5cm}
\label{table:hindi_english_comparisons}
\end{table}

%% file: tables/mkb-benchmark-M2M4.tex
\begin{table*}[h!]
    \centering
    {
        \begin{tabular}{|r|r|r|r|r|r|r|r|r|r|r|r|r|}
        \hline
&bn&en&gu&hi&ml&mr&or&ta&te&ur & $\Delta$\\ \hline
bn&&\cellcolor[rgb]{0.937870,0.964629,0.990158}{16.79}&\cellcolor[rgb]{0.922491,0.954787,0.985236}{15.50}&\cellcolor[rgb]{0.922491,0.954787,0.985236}{21.62}&\cellcolor[rgb]{0.953249,0.974471,0.995079}{5.75}&\cellcolor[rgb]{0.925567,0.956755,0.986221}{11.05}&\cellcolor[rgb]{0.790496,0.868174,0.941930}{12.42}&\cellcolor[rgb]{0.947097,0.970534,0.993110}{4.99}&\cellcolor[rgb]{0.937870,0.964629,0.990158}{5.64}&\cellcolor[rgb]{0.864314,0.917386,0.966536}{24.73}& 34.36
\\
en&\cellcolor[rgb]{0.947097,0.970534,0.993110}{8.74}&&\cellcolor[rgb]{0.937870,0.964629,0.990158}{12.92}&\cellcolor[rgb]{0.953249,0.974471,0.995079}{16.93}&\cellcolor[rgb]{0.953249,0.974471,0.995079}{5.51}&\cellcolor[rgb]{0.940946,0.966597,0.991142}{9.84}&\cellcolor[rgb]{0.837739,0.899669,0.957678}{9.07}&\cellcolor[rgb]{0.956324,0.976440,0.996063}{4.86}&\cellcolor[rgb]{0.937870,0.964629,0.990158}{5.75}&\cellcolor[rgb]{0.897885,0.939039,0.977363}{22.16}& 23.32
\\
gu&\cellcolor[rgb]{0.928643,0.958724,0.987205}{13.48}&\cellcolor[rgb]{0.934794,0.962661,0.989173}{21.93}&&\cellcolor[rgb]{0.907113,0.944944,0.980315}{44.16}&\cellcolor[rgb]{0.950173,0.972503,0.994095}{7.29}&\cellcolor[rgb]{0.916340,0.950850,0.983268}{17.22}&\cellcolor[rgb]{0.711265,0.831111,0.914433}{16.12}&\cellcolor[rgb]{0.947097,0.970534,0.993110}{6.06}&\cellcolor[rgb]{0.931719,0.960692,0.988189}{7.12}&\cellcolor[rgb]{0.787543,0.866205,0.940946}{45.82}& 45.70
\\
hi&\cellcolor[rgb]{0.925567,0.956755,0.986221}{13.84}&\cellcolor[rgb]{0.947097,0.970534,0.993110}{21.56}&\cellcolor[rgb]{0.882507,0.929196,0.972441}{35.79}&&\cellcolor[rgb]{0.950173,0.972503,0.994095}{7.75}&\cellcolor[rgb]{0.919416,0.952818,0.984252}{18.07}&\cellcolor[rgb]{0.711265,0.831111,0.914433}{16.40}&\cellcolor[rgb]{0.953249,0.974471,0.995079}{6.49}&\cellcolor[rgb]{0.934794,0.962661,0.989173}{7.69}&\cellcolor[rgb]{0.770319,0.856209,0.935102}{51.70}& 47.22
\\
ml&\cellcolor[rgb]{0.934794,0.962661,0.989173}{9.46}&\cellcolor[rgb]{0.937870,0.964629,0.990158}{17.01}&\cellcolor[rgb]{0.919416,0.952818,0.984252}{13.70}&\cellcolor[rgb]{0.922491,0.954787,0.985236}{20.02}&&\cellcolor[rgb]{0.925567,0.956755,0.986221}{10.78}&\cellcolor[rgb]{0.799354,0.874079,0.944883}{11.52}&\cellcolor[rgb]{0.947097,0.970534,0.993110}{5.93}&\cellcolor[rgb]{0.934794,0.962661,0.989173}{6.33}&\cellcolor[rgb]{0.852503,0.909512,0.962599}{23.88}& 36.30
\\
mr&\cellcolor[rgb]{0.931719,0.960692,0.988189}{11.34}&\cellcolor[rgb]{0.934794,0.962661,0.989173}{18.37}&\cellcolor[rgb]{0.919416,0.952818,0.984252}{19.53}&\cellcolor[rgb]{0.922491,0.954787,0.985236}{25.89}&\cellcolor[rgb]{0.944022,0.968566,0.992126}{6.56}&&\cellcolor[rgb]{0.784591,0.864237,0.939962}{12.95}&\cellcolor[rgb]{0.950173,0.972503,0.994095}{5.58}&\cellcolor[rgb]{0.934794,0.962661,0.989173}{6.21}&\cellcolor[rgb]{0.820023,0.887859,0.951772}{30.83}& 38.80
\\
or&\cellcolor[rgb]{0.882507,0.929196,0.972441}{12.98}&\cellcolor[rgb]{0.852503,0.909512,0.962599}{19.36}&\cellcolor[rgb]{0.817070,0.885890,0.950788}{19.94}&\cellcolor[rgb]{0.787543,0.866205,0.940946}{26.99}&\cellcolor[rgb]{0.925567,0.956755,0.986221}{6.21}&\cellcolor[rgb]{0.855456,0.911480,0.963583}{13.19}&&\cellcolor[rgb]{0.944022,0.968566,0.992126}{4.96}&\cellcolor[rgb]{0.907113,0.944944,0.980315}{5.65}&\cellcolor[rgb]{0.647290,0.803922,0.892042}{26.92}& 71.06
\\
ta&\cellcolor[rgb]{0.937870,0.964629,0.990158}{8.32}&\cellcolor[rgb]{0.940946,0.966597,0.991142}{15.30}&\cellcolor[rgb]{0.928643,0.958724,0.987205}{11.51}&\cellcolor[rgb]{0.928643,0.958724,0.987205}{17.20}&\cellcolor[rgb]{0.947097,0.970534,0.993110}{5.80}&\cellcolor[rgb]{0.928643,0.958724,0.987205}{9.26}&\cellcolor[rgb]{0.822976,0.889827,0.952757}{10.04}&&\cellcolor[rgb]{0.928643,0.958724,0.987205}{5.70}&\cellcolor[rgb]{0.855456,0.911480,0.963583}{20.56}& 33.31
\\
te&\cellcolor[rgb]{0.934794,0.962661,0.989173}{8.26}&\cellcolor[rgb]{0.928643,0.958724,0.987205}{12.92}&\cellcolor[rgb]{0.922491,0.954787,0.985236}{11.97}&\cellcolor[rgb]{0.919416,0.952818,0.984252}{17.47}&\cellcolor[rgb]{0.944022,0.968566,0.992126}{6.28}&\cellcolor[rgb]{0.928643,0.958724,0.987205}{9.53}&\cellcolor[rgb]{0.822976,0.889827,0.952757}{9.95}&\cellcolor[rgb]{0.940946,0.966597,0.991142}{5.53}&&\cellcolor[rgb]{0.828881,0.893764,0.954725}{21.99}& 36.71
\\
ur&\cellcolor[rgb]{0.922491,0.954787,0.985236}{12.92}&\cellcolor[rgb]{0.931719,0.960692,0.988189}{23.52}&\cellcolor[rgb]{0.882507,0.929196,0.972441}{28.76}&\cellcolor[rgb]{0.876355,0.925260,0.970473}{48.54}&\cellcolor[rgb]{0.934794,0.962661,0.989173}{7.66}&\cellcolor[rgb]{0.907113,0.944944,0.980315}{16.07}&\cellcolor[rgb]{0.799354,0.874079,0.944883}{11.60}&\cellcolor[rgb]{0.947097,0.970534,0.993110}{5.22}&\cellcolor[rgb]{0.931719,0.960692,0.988189}{6.85}&& 38.92
\\  \hline
$\Delta$ &24.12&24.16&38.96&38.92&13.47&31.19&109.16&11.74&22.32&91.66 & \\
\hline
        \end{tabular}
    }
    \caption{BLEU scores of M2M-4 model on multilingual test set Mann-Ki Baat. Rows correspond to source languages and
columns target languages. The colors indicate improvement (blue) or degradation (red) in comparison to M2M-1 \cite{siripragada-etal-2020-multilingual}. We observe cumulative increment of {+405} BLEU across all language pairs and a median increment of {+2.7}. The cumulative changes in translating to or from a given language in comparison to M2M-1 are provided under $\Delta$ header. It can be observed that related languages end up with higher BLEU scores without having to add the prior in the model formulation - e.g (hi, gu), (ur, gu), (ur, hi). Closely behind, there is (mr, hi) ahead of other language pairs.  %
}
    \label{table:mkb-benchmark}
        \vspacehackuniform{}

\end{table*}

%% file: appendix.tex
\clearpage

\section{Related Languages in Retrieval}

\label{para:related-languages}
\begin{figure}[h!]
\centering
\includegraphics[width=\linewidth]{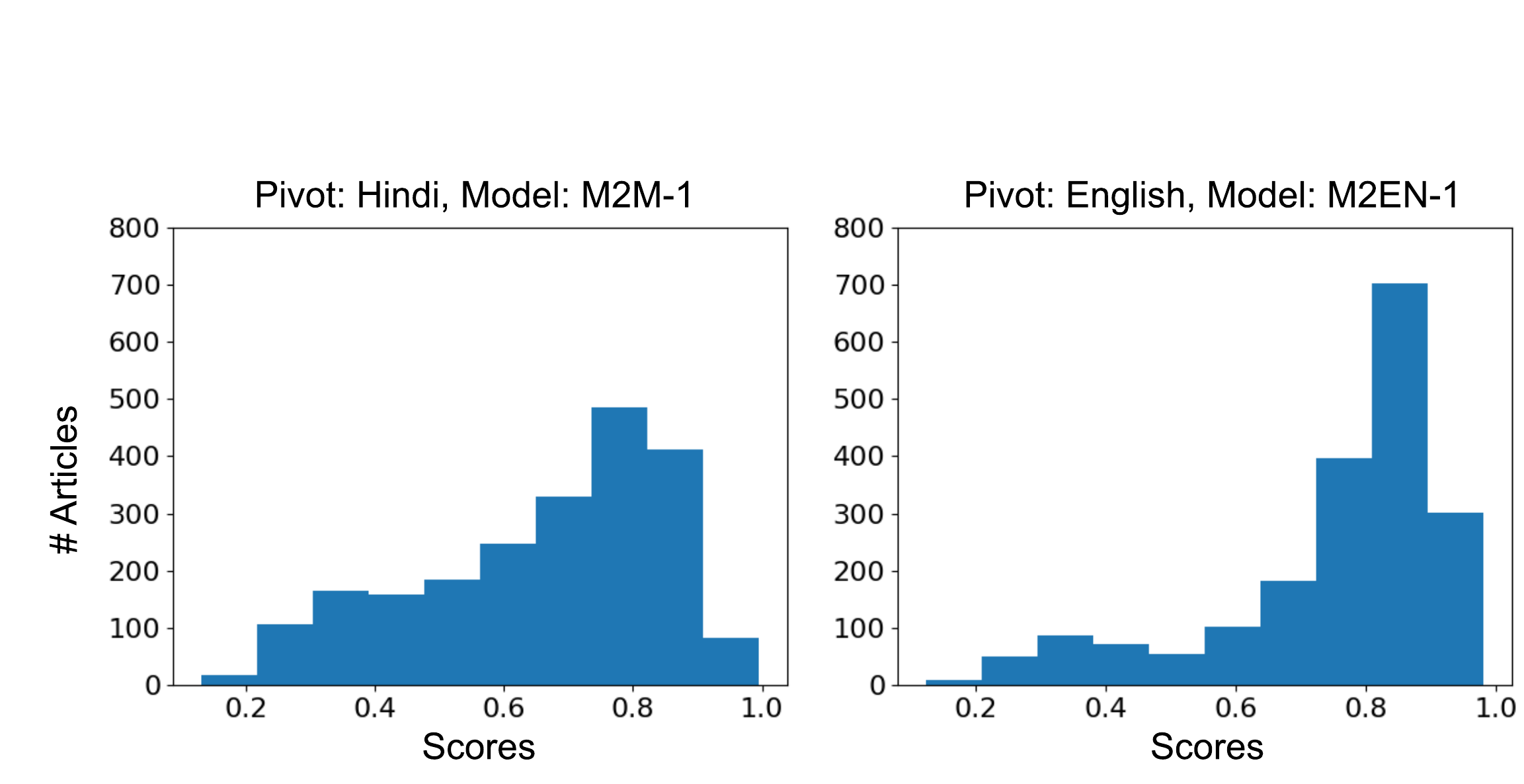}
\caption{Retrieval scores of \emph{Gujarati}. Left bar chart indicates retrieval scores in case of model M2M-1 and pivot language \emph{hi}. Right chart indicates scores in case of M2EN-1 and pivot \emph{en}.}
\label{pivot-lang-comparison}
\vspacehackuniform{}
\end{figure}

 The languages Marathi~(mr), Gujarati~(gu), Punjabi~(pa) are similar to Hindi and exhibit high BLEU scores (Table \ref{table:mkb-benchmark}) when translated to Hindi. They are also known to be similar~\cite{kunchukuttan2020utilizing}, so we experiment with hi as a pivot language.
However, we found poor retrieval performance when compared to pivoting through English. Upon closer inspection, we observed that Hindi articles are much more elaborate while describing content while the mr, gu, pa equivalents are often summarized. %
This is evident when considering examples of Gujarati articles, as PIB offices of Gujarat are responsible for posting the articles in Gujarati and their respective English translations. We illustrate this phenomenon through an example in Figure \ref{pivot-lang-comparison}, where we observe higher retrieval scores overall when compared to Hindi-based retrieval. The above analysis points to the success and a potential use-case of our model in being able to deliver consistent content across all languages for websites like PIB. %

\section{Implementation considerations}

The training was done on a machine equipped with 4 x NVIDIA 1080Tis or 2080Tis, 40 or 20 CPU cores and 128GB memory, depending on the allocation in our cluster. The multiparallel nature of the evolving PIB dataset and the ILCI dataset leads to an $O(N^2)$ growth with increase in samples across language pairs, specifically to train an M2M model. The training of the M2M models took \textasciitilde{}3 days and an M2EN model took \textasciitilde{}1 day from scratch, for the same number of epochs. Our hardware could not train the Transformer-Big \cite{vaswani2017attention} with the stability techniques prescribed by \cite{popel2018training}, so we had to use the Transformer-Base model.

We describe the code and modifications we implemented to train these models. We use fairseq \cite{ott2019fairseq} framework with some modifications\footnote{\href{https://github.com/jerinphilip/fairseq-ilmt}{https://github.com/jerinphilip/fairseq-ilmt}}. Our modifications for this work include a dataloader equipped with memory-mapped storage using LMDB for fast access from the corpora described in Table \ref{table:iter0-dataset-stats}. The non-standard SentencePiece routine required some additional integrations, and these are publicly available as well\footnote{\href{https://github.com/jerinphilip/ilmulti}{https://github.com/jerinphilip/ilmulti}}. We provide wrappers for using our trained models for inference in Python packaging, with models available for download separately.

\begin{table}[h]
    \centering
    \begin{tabular}{cccc}
    \toprule
        Device  & Sentences &  Batched & Time   \\
        \midrule
        CPU  & 100  & Yes & 24.08\\
        GPU & 100 &  Yes & 2.06 \\
        CPU   & 100 & No & 60.84 \\
        GPU & 100 & No & 30.69 \\
    \bottomrule
    \end{tabular}
    \caption{Caption}
    \label{table:hardware-benchmarks}
\end{table}

We benchmarked the inference pipeline on both CPU and GPU machines. We present the summary of time taken to translate a sample test-set of sentences in Table \ref{table:hardware-benchmarks}. The inference can be done on a CPU for some practical use-cases, and our hosted demo model\footnote{\href{http://preon.iiit.ac.in/babel/gui}{http://preon.iiit.ac.in/babel/gui}} runs on a CPU.

\input{tables/ilci-benchmark-kunchukuttan-M2M3}

\section{ILCI numbers}

In Table \ref{table:ilci-benchmark}, we provide comparisons with ILCI, with \citet{kunchukuttan2014sata}. Despite not being domain adapted to ILCI, we obtain better BLEU scores in a majority of language-directions. 

ILCI has been used by several works in the past with non-standard or non-reproducible performance benchmarks, which has rendered comparison hard with these. We urge the community to avoid using splits in ILCI for publishing results, absent any method to reproduce the constituent sentences in the split.

\section{Versioning}

Table \ref{table:pib-v0.2-grid} provides the sizes of CVIT PIBv0.2, the corpus used in WMT-2020 and WAT-2020. The corpus is generated with M2EN-2 using articles crawled from PIB posted until December 2019.

CVIT PIBv1.3 contains articles crawled until August 2020. Future releases will be described on the project website\footnote{\url{http://preon.iiit.ac.in/~jerin/bhasha} page.}.

\input{tables/pib-final-grid}

%% file: tables/ilci-benchmark-kunchukuttan-M2M3.tex
\begin{table*}[h!]
    \centering
    {
        \begin{tabular}{|c|c|c|c|c|c|c|c|c|c|c|c|}
        \hline
&bn&en&gu&hi&ml&mr&pa&ta&te&ur \\
bn&&\cellcolor[rgb]{0.900961,0.941007,0.978347}{22.13}&\cellcolor[rgb]{0.996432,0.885859,0.834141}{26.34}&\cellcolor[rgb]{0.996801,0.893610,0.845213}{31.72}&\cellcolor[rgb]{0.999754,0.955617,0.933795}{8.28}&\cellcolor[rgb]{0.999139,0.942699,0.915340}{18.62}&\cellcolor[rgb]{0.993341,0.827789,0.756463}{24.45}&\cellcolor[rgb]{0.998647,0.932364,0.900577}{7.12}&\cellcolor[rgb]{0.965552,0.982345,0.999016}{13.19}&\cellcolor[rgb]{0.947097,0.970534,0.993110}{26.10}\\
en&\cellcolor[rgb]{0.962476,0.980377,0.998032}{15.13}&&\cellcolor[rgb]{0.837739,0.899669,0.957678}{24.82}&\cellcolor[rgb]{0.897885,0.939039,0.977363}{30.42}&\cellcolor[rgb]{0.944022,0.968566,0.992126}{6.10}&\cellcolor[rgb]{0.846597,0.905575,0.960631}{17.12}&\cellcolor[rgb]{0.962476,0.980377,0.998032}{23.05}&\cellcolor[rgb]{0.944022,0.968566,0.992126}{5.52}&\cellcolor[rgb]{0.894810,0.937070,0.976378}{10.44}&\cellcolor[rgb]{0.858408,0.913449,0.964567}{24.29}\\
gu&\cellcolor[rgb]{0.994571,0.850550,0.786605}{22.91}&\cellcolor[rgb]{0.781638,0.862268,0.938977}{30.32}&&\cellcolor[rgb]{0.944022,0.968566,0.992126}{54.47}&\cellcolor[rgb]{0.950173,0.972503,0.994095}{8.88}&\cellcolor[rgb]{0.944022,0.968566,0.992126}{28.38}&\cellcolor[rgb]{0.996048,0.877862,0.822776}{42.12}&\cellcolor[rgb]{0.998770,0.934948,0.904268}{8.38}&\cellcolor[rgb]{0.999262,0.945283,0.919031}{15.64}&\cellcolor[rgb]{0.778685,0.860300,0.937993}{45.56}\\
hi&\cellcolor[rgb]{0.988235,0.701807,0.595233}{24.26}&\cellcolor[rgb]{0.913264,0.948881,0.982284}{31.19}&\cellcolor[rgb]{0.999877,0.958201,0.937486}{53.20}&&\cellcolor[rgb]{0.998893,0.937532,0.907958}{9.44}&\cellcolor[rgb]{0.998770,0.934948,0.904268}{32.06}&\cellcolor[rgb]{0.988235,0.600923,0.479585}{54.55}&\cellcolor[rgb]{0.998647,0.932364,0.900577}{9.61}&\cellcolor[rgb]{0.997170,0.901361,0.856286}{17.67}&\cellcolor[rgb]{0.796401,0.872111,0.943899}{60.13}\\
ml&\cellcolor[rgb]{0.891734,0.935102,0.975394}{14.34}&\cellcolor[rgb]{0.799354,0.874079,0.944883}{18.01}&\cellcolor[rgb]{0.808212,0.879985,0.947835}{20.13}&\cellcolor[rgb]{0.793449,0.870142,0.942914}{24.49}&&\cellcolor[rgb]{0.843645,0.903606,0.959646}{14.04}&\cellcolor[rgb]{0.855456,0.911480,0.963583}{18.88}&\cellcolor[rgb]{0.934794,0.962661,0.989173}{6.41}&\cellcolor[rgb]{0.885582,0.931165,0.973426}{10.82}&\cellcolor[rgb]{0.781638,0.862268,0.938977}{20.88}\\
mr&\cellcolor[rgb]{0.997785,0.914279,0.874740}{20.70}&\cellcolor[rgb]{0.775240,0.858301,0.936824}{26.83}&\cellcolor[rgb]{0.910188,0.946913,0.981300}{37.12}&\cellcolor[rgb]{0.931719,0.960692,0.988189}{43.62}&\cellcolor[rgb]{0.944022,0.968566,0.992126}{8.50}&&\cellcolor[rgb]{0.998647,0.932364,0.900577}{33.00}&\cellcolor[rgb]{0.999262,0.945283,0.919031}{7.47}&\cellcolor[rgb]{0.928643,0.958724,0.987205}{14.11}&\cellcolor[rgb]{0.790496,0.868174,0.941930}{35.20}\\
pa&\cellcolor[rgb]{0.992603,0.814133,0.738378}{22.42}&\cellcolor[rgb]{0.837739,0.899669,0.957678}{31.47}&\cellcolor[rgb]{0.913264,0.948881,0.982284}{49.28}&\cellcolor[rgb]{0.997908,0.916863,0.878431}{68.46}&\cellcolor[rgb]{0.937870,0.964629,0.990158}{9.21}&\cellcolor[rgb]{0.904037,0.942976,0.979331}{29.17}&&\cellcolor[rgb]{0.999877,0.958201,0.937486}{8.92}&\cellcolor[rgb]{0.998770,0.934948,0.904268}{16.27}&\cellcolor[rgb]{0.784591,0.864237,0.939962}{54.94}\\
ta&\cellcolor[rgb]{0.997047,0.898777,0.852595}{10.77}&\cellcolor[rgb]{0.907113,0.944944,0.980315}{14.36}&\cellcolor[rgb]{0.999385,0.947866,0.922722}{16.53}&\cellcolor[rgb]{0.999262,0.945283,0.919031}{20.82}&\cellcolor[rgb]{0.999508,0.950450,0.926413}{5.76}&\cellcolor[rgb]{0.965552,0.982345,0.999016}{11.22}&\cellcolor[rgb]{0.998032,0.919446,0.882122}{16.63}&&\cellcolor[rgb]{0.999385,0.947866,0.922722}{8.60}&\cellcolor[rgb]{0.931719,0.960692,0.988189}{17.65}\\
te&\cellcolor[rgb]{0.999385,0.947866,0.922722}{16.02}&\cellcolor[rgb]{0.811165,0.881953,0.948820}{21.04}&\cellcolor[rgb]{0.900961,0.941007,0.978347}{25.90}&\cellcolor[rgb]{0.919416,0.952818,0.984252}{29.90}&\cellcolor[rgb]{0.956324,0.976440,0.996063}{7.12}&\cellcolor[rgb]{0.913264,0.948881,0.982284}{16.44}&\cellcolor[rgb]{0.998647,0.932364,0.900577}{23.34}&\cellcolor[rgb]{0.999631,0.953033,0.930104}{6.80}&&\cellcolor[rgb]{0.861361,0.915417,0.965552}{25.01}\\
ur&\cellcolor[rgb]{0.994817,0.855102,0.792634}{19.80}&\cellcolor[rgb]{0.840692,0.901638,0.958662}{28.18}&\cellcolor[rgb]{0.900961,0.941007,0.978347}{42.60}&\cellcolor[rgb]{0.998770,0.934948,0.904268}{56.41}&\cellcolor[rgb]{0.947097,0.970534,0.993110}{8.54}&\cellcolor[rgb]{0.913264,0.948881,0.982284}{24.33}&\cellcolor[rgb]{0.991619,0.795925,0.714264}{43.23}&\cellcolor[rgb]{0.956324,0.976440,0.996063}{8.70}&\cellcolor[rgb]{0.999754,0.955617,0.933795}{14.39}&\\
\hline
        \end{tabular}
    }
    \caption{BLEU scores of inference of model M2M-3 on random test split from ILCI. 
The reds indicate poorer performance compared to \citet{kunchukuttan2014sata} and the blues better performance. Overall, our model performs better in 52/90 tasks and is +121 BLEU points ahead of \emph{Sata Anuvaadak} with a median BLEU increase of 1.1. }
    \label{table:ilci-benchmark}
\end{table*}

%% file: tables/pib-final-grid.tex
\newcolumntype{Y}{>{\small\raggedleft\arraybackslash}p{0.7cm}}

\begin{table*}[h]
\centering
{
\small 
    \begin{tabular}{YYYYYYYYYYYY}
    \toprule
&en&hi&ta&te&ml&ur&bn&gu&mr&or&pa \\ \midrule
en&&\cellcolor[rgb]{0.740792,0.843660,0.924767}{195208}&\cellcolor[rgb]{0.822976,0.889827,0.952757}{87113}&\cellcolor[rgb]{0.999877,0.958201,0.937486}{5752}&\cellcolor[rgb]{0.885582,0.931165,0.973426}{31974}&\cellcolor[rgb]{0.999877,0.958201,0.937486}{45344}&\cellcolor[rgb]{0.820023,0.887859,0.951772}{48354}&\cellcolor[rgb]{0.947097,0.970534,0.993110}{29421}&\cellcolor[rgb]{0.726028,0.837386,0.919600}{80760}&\cellcolor[rgb]{0.647290,0.803922,0.892042}{58461}&\cellcolor[rgb]{0.962476,0.980377,0.998032}{27117}\\
hi&&&\cellcolor[rgb]{0.867266,0.919354,0.967520}{44031}&\cellcolor[rgb]{0.959400,0.978408,0.997047}{3083}&\cellcolor[rgb]{0.913264,0.948881,0.982284}{17819}&\cellcolor[rgb]{0.956324,0.976440,0.996063}{11695}&\cellcolor[rgb]{0.888658,0.933133,0.974410}{24849}&\cellcolor[rgb]{0.931719,0.960692,0.988189}{19730}&\cellcolor[rgb]{0.837739,0.899669,0.957678}{45950}&\cellcolor[rgb]{0.781638,0.862268,0.938977}{36317}&\cellcolor[rgb]{0.947097,0.970534,0.993110}{11442}\\
ta&&&&\cellcolor[rgb]{0.962476,0.980377,0.998032}{3218}&\cellcolor[rgb]{0.913264,0.948881,0.982284}{15029}&\cellcolor[rgb]{0.959400,0.978408,0.997047}{4964}&\cellcolor[rgb]{0.897885,0.939039,0.977363}{19175}&\cellcolor[rgb]{0.919416,0.952818,0.984252}{16934}&\cellcolor[rgb]{0.849550,0.907543,0.961615}{33636}&\cellcolor[rgb]{0.820023,0.887859,0.951772}{27668}&\cellcolor[rgb]{0.944022,0.968566,0.992126}{9150}\\
te&&&&&\cellcolor[rgb]{0.959400,0.978408,0.997047}{2543}&\cellcolor[rgb]{0.999877,0.958201,0.937486}{415}&\cellcolor[rgb]{0.962476,0.980377,0.998032}{1883}&\cellcolor[rgb]{0.962476,0.980377,0.998032}{2625}&\cellcolor[rgb]{0.956324,0.976440,0.996063}{2627}&\cellcolor[rgb]{0.959400,0.978408,0.997047}{1834}&\cellcolor[rgb]{0.965552,0.982345,0.999016}{1220}\\
ml&&&&&&\cellcolor[rgb]{0.959400,0.978408,0.997047}{2378}&\cellcolor[rgb]{0.928643,0.958724,0.987205}{9940}&\cellcolor[rgb]{0.934794,0.962661,0.989173}{10132}&\cellcolor[rgb]{0.907113,0.944944,0.980315}{14474}&\cellcolor[rgb]{0.913264,0.948881,0.982284}{9843}&\cellcolor[rgb]{0.950173,0.972503,0.994095}{4961}\\
ur&&&&&&&\cellcolor[rgb]{0.956324,0.976440,0.996063}{3795}&\cellcolor[rgb]{0.965552,0.982345,0.999016}{2397}&\cellcolor[rgb]{0.947097,0.970534,0.993110}{4941}&\cellcolor[rgb]{0.953249,0.974471,0.995079}{3209}&\cellcolor[rgb]{0.965552,0.982345,0.999016}{5584}\\
bn&&&&&&&&\cellcolor[rgb]{0.934794,0.962661,0.989173}{10554}&\cellcolor[rgb]{0.882507,0.929196,0.972441}{19914}&\cellcolor[rgb]{0.888658,0.933133,0.974410}{14606}&\cellcolor[rgb]{0.947097,0.970534,0.993110}{5332}\\
gu&&&&&&&&&\cellcolor[rgb]{0.904037,0.942976,0.979331}{17169}&\cellcolor[rgb]{0.894810,0.937070,0.976378}{13682}&\cellcolor[rgb]{0.956324,0.976440,0.996063}{5581}\\
mr&&&&&&&&&&\cellcolor[rgb]{0.799354,0.874079,0.944883}{31377}&\cellcolor[rgb]{0.931719,0.960692,0.988189}{9601}\\
or&&&&&&&&&&&\cellcolor[rgb]{0.934794,0.962661,0.989173}{6813}\\

\bottomrule
    \end{tabular}
}
\caption{Multilingual shared content across language pairs for CVIT-PIBv0.2. Rows and columns indicate language pairs. The highlights are proportional to the change after the iterative alignment process, reds indicating decrease and blues indicating increases in corpora sizes compared to the previous release v0.0. Evident from the table, we notice major increments in Marathi, Oriya and other languages. Tamil and Hindi which we had enough to be considered mid-to-high resource gain significant number as well. The maximum decrease is -283 for Telugu, which is negligible compared to the improvements of the order of ten-thousands in many language-pairs.}
\label{table:pib-v0.2-grid}
    \vspacehackuniform{}
\end{table*}